\def\year{2019}\relax
\documentclass[letterpaper]{article} 
\usepackage{aaai19}  
\usepackage{times}  
\usepackage{helvet}  
\usepackage{courier}  
\usepackage{url}  
\usepackage{graphicx}  
\usepackage{amsmath}
\usepackage{booktabs}
\usepackage{amsfonts}
\usepackage{amssymb}
\usepackage{listings}
\usepackage{subcaption}
\usepackage[T1]{fontenc}
\usepackage{hyperref}

\usepackage{color}

\definecolor{teal}{RGB}{0, 173, 188}
\definecolor{purple}{RGB}{118, 101, 160}
\definecolor{gold}{RGB}{255, 164, 0}
\definecolor{gray}{RGB}{102, 102, 102}

\begin{document}
%

\title{Zero Shot Learning for Code Education:\\ Rubric Sampling with Deep Learning Inference}
\author{
    Mike Wu\textsuperscript{1}, Milan Mosse\textsuperscript{1}, Noah Goodman\textsuperscript{1,2}, Chris Piech\textsuperscript{1}\\
    \textsuperscript{1} Department of Computer Science, Stanford University, Stanford, CA 94305\\
    \textsuperscript{2} Department of Psychology, Stanford University, Stanford, CA 94305\\
    \texttt{\{wumike,mmosse19,ngoodman,piech\}@stanford.edu}
}
\maketitle
\begin{abstract}
In modern computer science education, massive open online courses (MOOCs) log thousands of hours of data about how students solve coding challenges. Being so rich in data, these platforms have garnered the interest of the machine learning community, with many new algorithms attempting to autonomously provide feedback to help future students learn. But what about those \textit{first} hundred thousand students? In most educational contexts (i.e. classrooms), assignments do not have enough historical data for supervised learning. In this paper, we introduce a human-in-the-loop ``rubric sampling" approach to tackle the ``zero shot" feedback challenge. We are able to provide autonomous feedback for the first students working on an introductory programming assignment with accuracy that substantially outperforms data-hungry algorithms and approaches human level fidelity. Rubric sampling requires minimal teacher effort, can associate feedback with specific parts of a student's solution and can articulate a student's misconceptions in the language of the instructor. Deep learning inference enables rubric sampling to further improve as more assignment specific student data is acquired. We demonstrate our results on a novel dataset from Code.org, the world's largest programming education platform.
\end{abstract}

\section{Introduction}
\label{sec:introduction}

The need for high quality education at scale poses a difficult challenge. The price of education per student is growing faster than economy-wide costs \cite{bowen2012cost}, limiting the resources available to support student learning. When also considering the rising need to provide adult retraining, the gap between the demand for education and our ability to provide is especially large. In recent years, massively open online courses (MOOCs) from platforms like Coursera and Code.org have made progress by scaling the delivery of \emph{content}. However, MOOCs largely ignore an equally important ingredient for learning: high quality \textit{feedback}. The clear societal need, alongside massive amounts of data has led to a machine learning grand challenge: learn how to provide feedback for education at scale, especially in computer science due to its apparent structure and high demand.

Scaling feedback (a.k.a. ``feedback" challenge) has proven to be a hard machine learning problem. Despite dozens of projects to combine massive datasets with cutting edge deep learning, current approaches fall short. Three issues emerge: (1) for even basic computer science education, homework datasets have statistical distributions with heavy tails similar to natural language; (2) hand labeling feedback is expensive, rendering supervised solutions infeasible; (3) in real world contexts feedback is needed for assignments with small (or zero) historical records of student learning. For the billions of learners around the world, most education and assessments have \textit{at most} hundreds of records. Even if students use Code.org, assignments are constantly changing, making the small-data context  perennial. It is a zero-shot solution that has potential to deliver enormous social impact.

We build upon a simple insight that enables us to move beyond the supervised  paradigm:
When experts give feedback, they are asked to perform the hard task of predicting misconception ($y$) given program ($x$). When breaking down the cognitive steps that experts go through, they often solve the inference by first thinking generatively $p(x, y )$. They imagine, ``if a student were to have a particular set of misconceptions, what sorts of programs is he or she likely to produce." Thinking generatively is much easier: while there are a finite set of decomposable misconceptions, they combine into exponential amounts of unique solutions.

We formalize this intuition into a technique we call ``rubric sampling" to elicit samples from an expert prior of the joint distribution $p(x, y)$ and use deep learning for inference $p(y|x)$. With no historical examples, rubric sampling enables feedback with accuracy close to the fidelity of human teachers, outperforming data-intensive state of the art algorithms. We case study this  technique on  Code.org, an online programming platform that has been used by 610 million students and has provided a full curriculum to 29 million students, equivalent to 39\% of the US K12 population.

\paragraph{Specific contributions in this paper:}

\begin{enumerate}
    \item We introduce the Zero Shot Feedback Challenge on a dataset from 8 assignments from Code.org along with an evaluation set of 800 labels.
    \item We articulate a novel solution: rubric sampling with deep learning inference which sets the new state of the art in code feedback prediction: F1 score doubled over baseline, approaching human level accuracy.
    \item We introduce the ability to (i) attribute feedback to specific parts of code, (ii) trace learning over time and (iii) generate synthetic datasets.
\end{enumerate}

\begin{figure*}[h!]
\centering
\includegraphics[width=\textwidth]{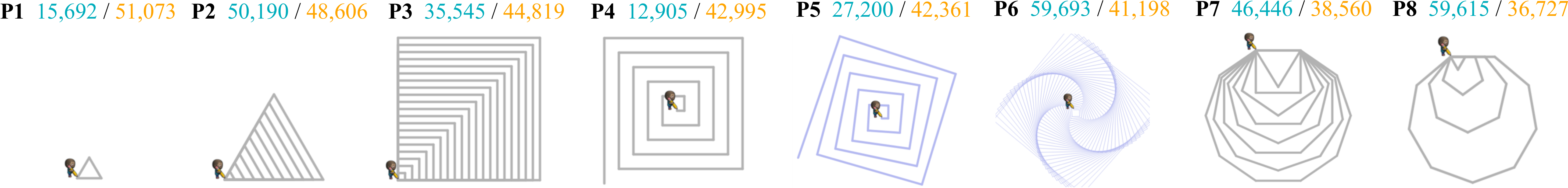}
\caption{The curricula for learning nested for loops in Code.org. To provide intuition on the vast domain complexity, we show the \textcolor{teal}{number of unique solutions} and the \textcolor{gold}{number of students} who attempted the problem for each of the 8 exercises.}
\label{fig:policy}
\end{figure*}

\section{The Zero Shot Feedback Challenge}


The ``Zero-Shot" Feedback Challenge is to infer the misconceptions that led to errors in a student's answer using zero historical examples of student work and zero expert annotations. Though this challenge is difficult, it is a task that humans find straightforward. Experts are especially adept at generalizing: an instructor does not need to see thousands of instances of a misunderstanding in order to understand it.

\textit{Why is zero-shot so important?} Human annotated examples are surprisingly time consuming to acquire. In 2014, Code.org launched an initiative where hundreds of thousands of instructors were crowdsourced to provide feedback to student solutions\footnote{\url{http://www.code.org/hints}}. Labeling was hard and undesirable work and the long tail of unique solutions meant that even after thousands of human hours of teacher work, the annotations were only scratching the surface of feedback. The initiative was cancelled after two years and the effort has not been reproduced since. For small classrooms and massive online platforms alike, it is infeasible to acquire the supervision required for contemporary nonlinear methods.

We foresee three main approaches: (1) learn to transfer information from one assignment to another, (2) learn to incorporate expert knowledge, and (3) form algorithms that can generalize from small amounts of human annotations.

\label{sec:challenge}

\subsection{Related Work}

\paragraph{Education Feedback}

If you were to solve an assignment on Code.org today, the hints you would be given are generated from a unit test system combined with static analysis of the students solution. It has been a widely reported social-good objective to improve upon these hints \cite{price2017position} especially since the state of the art is far from ideal \cite{o2014hint}. Achieving this goal has proved to be hard. Previous research on a more basic set of Code.org challenges (the ``Hour of Code") have scratched the surface with respect to providing feedback at scale. Original work  found that latent patterns in \emph{how} students solve programming assignments have signal as to how he or she should proceed \cite{piech2015autonomously}. Applying a neural network improved prediction of feedback \cite{programEmbeddings} but models were (1) too far from human accuracy, (2) weren't able to explain its predictions and (3) required massive amounts of data. The current state of the art combines these ideas and provides some improvements \cite{wang2017learning}.  In this paper we propose a method which uses less data, approaches human accuracy and works on more complex Code.org assignments by diverging from the classic supervised framework. Research on feedback for even more complex assignments such as medical education \cite{geigle2016exploration} and natural language questions \cite{bulgarov2018proposition} has also relied on data-hungry supervised learning and perhaps would  benefit from a rubric sampling inspired approach.

Theoretical inspiration for our expert-based generative rubric sampling derives from Brown's ``Repair Theory" which argues that the best way to help students is to understand the generative origins of their mistakes  \cite{brown1980repair}. Simulating student cognition has been applied to simple arithmetic problems \cite{koedinger2015methods} and recent hard coded models have been very successful in inferring why students make subtraction mistakes \cite{feldman2018automatic}. Researchers have argued that such expert models are infeasible for assignments as complex as coding \cite{paassen2017continuous}. However, the automated hierarchical decomposition achieved by \cite{nguyen2014codewebs} inspired us to develop rubric sampling, a simple expert model that works for programming assignments.

\paragraph{Zero Shot Learning}

There is a growing body of work in zero shot learning from the machine learning community, spawned by poor performance on unseen data.

The simplest approach is to include a human-in-the-loop. While labeling is one way human experts can ``teach" a model, it is often not the most efficient. Instead, \cite{lee2017multi} leverages knowledge graphs build by humans to estimate a similarity score. Similarly, \cite{lake2015human} present a  probabilistic knowledge graph (i.e. a Bayesian program) for generating symbols that outperform deep learning on out-of-sample alphabets. In this work, we employ a specific knowledge graph called a grammar, which we find to improve generalization.

A more complex approach (without human involvement) focuses on unsupervised algorithms to estimate the data distribution. \cite{verma2018generalized,wang2017zero} train an adversarial autoencoder by generating \textit{synthetic} examples and concurrently fitting a discriminator to classify between synthetic and empirical data. \cite{xian2018feature} propose a similar method for a CNN feature space. In this paper, we generalize this technique to a larger class of (non-differentiable) models: in lieu of a discriminator, we interpolate between synthetic and empirical data via a multimodal autoencoder.

\subsection{The Code.org Exercises Dataset}

\textit{Code.org} is an online education platform for teaching beginners fundamental concepts in programming. Students build their solutions in a drag-and-drop interface that pieces together blocks of code. Its growing popularity since 2013 has captured a large audience, having been used by 610 million students worldwide. We investigate a single curriculum consisting of 8 exercises from Code.org's catalog. In this particular unit, students are learning to combine nested for loops with variables, and particularly the use of a for loop counter in calculations. The problems are all presented as drawing geometric shapes in a 2D coordinate space, requiring knowledge of angles. For instance, the curriculum begins with the task of drawing an equilateral triangle (see Figure~\ref{fig:policy}).

\begin{figure}[h!]
    \centering
    \includegraphics[width=0.45\textwidth]{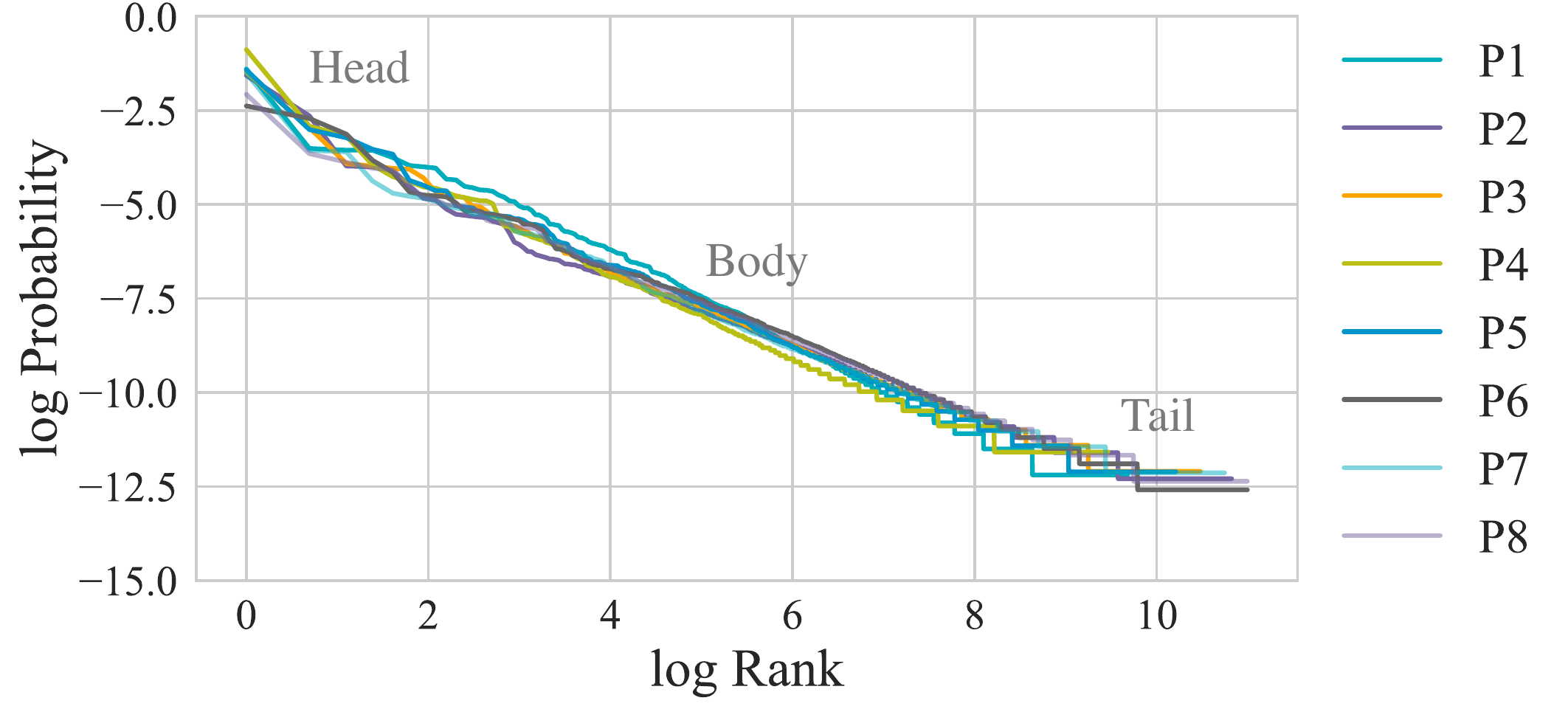}
    \caption{The distribution of programs for 8 problems from Code.org follow closely to a Zipf distribution, as shown by the linear relationship between the log probability of a program and the log of its rank in frequency. 5 to 10 programs dominate while the rest are in the heavy tails.}
    \label{fig:zipfs}
\end{figure}

The dataset is compiled from 54,488 students. Each time a student runs their code, the submission is saved. Each student is then associated with a \textit{trajectory} of programs whose length depends on how many tries the student took. In total, there are 1,598,375 submissions. Since the exercises do not have a bounded solution space, students could produce arbitrarily long programs. This implies that, much like natural language, the distribution of student submissions has extremely heavy tails. Figure~\ref{fig:zipfs} shows how closely the submissions follow a Zipf distribution. To emphasize the difficulty, even after a million students, there is still a 15\% chance that a new student generates a solution never seen before.

\paragraph{Evaluation Metric}

If we only cared about accuracy, we would prioritize the handful of 5 to 10 programs that make up the majority of the dataset. But given the infinite number of possible programs, struggling students who would benefit most from feedback will almost certainly not submit any one of the ``most likely" programs. Knowing this, we define our evaluation metrics in terms of the Zipf itself: let the \textit{head} (of the Zipf) refer to the top 20 programs ordered by frequency, the \textit{tail} as any program with a frequency of three or less, and the \textit{body} as everything in between. Figure~\ref{fig:zipfs} shows the rough placement of these three splits. When evaluating models, we ignore the head: these \textit{very common} programs can be manually labeled. Instead, we will report two F1 scores\footnote{We choose F1 score over accuracy as the labels are not close to balanced. Thus, accuracy tends to overinflate numbers.}: one for programs in the body and one for the tail.

\begin{figure*}[t!]
    \centering
    \includegraphics[width=\textwidth]{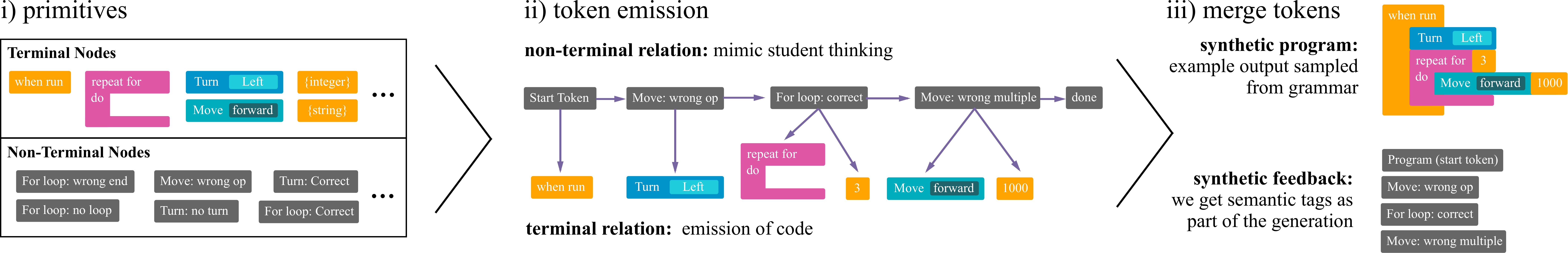}
    \caption{\textit{Probabilistic grammar (PCFG) for synthetic block-based programs.} To generate a synthetic example, we sequentially choose a set of non-terminal nodes, each of which will emit a set of terminal nodes. The composition of these terminal nodes make up a program whereas the composition of non-terminal nodes make up the labels/feedback. The emission and transition probabilities are specified by a human designer (or learned via evolutionary strategies).}
    \label{fig:grammar_diagram}
\end{figure*}

\paragraph{Human Annotations}
We collected fine-grained human annotations to over 800 \textit{unique} solutions (chosen randomly from P1 and P8) from 7 teaching assistants from the Stanford introduction to programming course. We chose to label programs from the easiest (P1) and hardest (P8) exercises as they are most different. Intermediate exercises (P2 to P7) share many of same structural features. The annotations are binary labels of 20 misconceptions that cover geometric concepts (e.g. doesn't understand equilateral is 60 degrees) to control flow concepts (e.g. repeats code instead of using a loop). 200 annotations were used to measure inter-rater reliability and the remaining 600 were used for evaluation. We refer to this  dataset as $D$. Labeling took 25.9 hours (117 seconds per program). At this rate, the entire dataset would take 9987 hours of expert time, over a year of continual work.

\section{Methods}
We consider learning tasks given a dataset of $n$ labeled examples, where each example (indexed by $i$) has an input string $x_i$ and a target output vector $y_i = [y_{i,1}, ..., y_{i,l}]$ composed of $l$ independent binary labels. In this context, we assume each string represents a block-based program in Lisp-like notation. Specifically, a program string is a sequence of $T_i$ tokens, each representing either an operator (functions, \textit{for} loop, \textit{if} statements, etc.), an operand (i.e. variable), an open parenthesis ``(", or a close parenthesis ``)". See Listing~\ref{lst:example} for an example program. Formally then, we describe the dataset as: $D = \{x_i, y_i\}_{i=1}^{n}$ where $x_i = [x_{i,1}, ..., x_{i, T_i}]$. The goal is to learn a function $\hat{y}_i = f(x_i)$ such that we minimize the error metric defined above, $err(\hat{y}_i, y_i)$. For supervised approaches, we split $D$ into a training ($D_\text{train}$) and test set ($D_\text{test}$) via a 80-20 ratio. For unsupervised methods, we evaluate on the entire set $D$.

\begin{lstlisting}[breaklines=true,caption={Example $x_i$ from P1 with 51 tokens. The tokens \textit{Program} and \textit{WhenRun} serve the role of a start-of-sentence tokens. A model will receive each token in order.},label={lst:example}]
( Program ( WhenRun ) ( Move ( Forward ) ( Value ( Number ( 50 ) ) ) ) ( Repeat ( Value ( Number ( 3 ) ) ) ( Body ( Turn ( Left ) ( Value ( Number ( 120 ) ) ) ) ) ) )
\end{lstlisting}

\subsection{Baselines}

\paragraph{Majority Label} As a sanity check, we can naively make predictions by voting for the majority label from $D_\text{train}$.

\paragraph{Predicting from Output} The ubiquitous way to provide feedback is via unit tests: analyze the output of executing $x_i$. For Code.org, each execution trace results in a sequence of output vectors $o_i = (o_{i,1}, o_{i,2}, ...)$ representing coordinates in 2D space where a line has been drawn. We train a recurrent neural network (RNN) to predict $y_i$ from $o_i$. Unfortunately, this model requires $x_i$ to compile.

\paragraph{Feedforward Neural Network} To circumvent compilation, one can tackle the more difficult problem of predicting feedback from raw program strings \cite{piech2015learning}. We train a $l$-dimensional classifier composed of a RNN over tokens by minimizing the binary cross entropy between predictions $\hat{y}_i$ and ground truth $y_i$ vectors.
\begin{equation}
    \min \sum_{j=1}^{l} [-(y_{i,j} \log \hat{y}_{i,j}) + (1 - y_{i,j}) \log(1 - \hat{y}_{i,j})]
\end{equation}
The model architecture borrows the sentence encoder (without any stochastic layers) from \cite{bowman2015generating} and concatenates a 3-layer perceptron with a softmax over $l$ output dimensions. As we will reuse these architectures for other models, we refer to deterministic encoder as the \textit{program network} and the latter MLP as the \textit{feedback network}.

\paragraph{Trajectory Prediction} No model so far uses the fact that each student submits many programs before either stopping or reaching the correct solution. In fact, the current SOTA \cite{wang2017learning} is to associate a \textit{trajectory} of $k$ programs $(x_1, x_2, ..., x_k)$ with the label $y_i$ corresponding to the last program, $x_k$. Then, for each program $x_i$, we train an embedding $e_i = f(z|x_i)$, where $f$ is the program network. This results in a sequence of embeddings $(e_1, e_2, ..., e_k)$ for a single trajectory. We concurrently train a \textit{second} RNN to compress the sequence to a single vector $z_i = \text{RNN}(e_1, e_2, ..., e_k)$. This is then provided as input to the feedback network. The hope is that structure induced by a trajectory implicitly provides labels that strengthen learning.

\paragraph{Deep Generative Model} Finally, we present a new baseline that is promising in the context of limited data. If we consider programs and feedback as two modalities, one  approach is to capture the joint distribution $p(x_i, y_i)$. Doing so, we can make predictions by sampling from the conditional having seen the program:  $\hat{y}_i \sim p(y_i|x_i)$. To do this, we train a multimodal variational autoencoder, MVAE \cite{wu2018multimodal} with two channels. Succinctly, this generative model uses a product-of-experts inference network where the joint distribution factorizes into a product of distributions defined by two modalities: $q(z|x, y) = q(z|x)q(z|y)$ where $x$ and $y$ are two observed modalities and $z$ is a latent variable. We optimize the multimodal evidence lower bound \cite{wu2018multimodal,vedantam2017generative}, which is a sum of three lower bounds:

\begin{align}
    \small
    & \mathop{\mathbb{E}}_{q_{\phi_h}(z|x,y)}[\sum_{h\in\{x,y\}}\lambda_h\log p_{\theta_h}(h|z)] - \beta \text{KL}[q_{\phi_h}(z|x,y),p(z)]  \nonumber \\
    & + \mathop{\mathbb{E}}_{q_{\phi_x}(z|x)}[\log p_{\theta_x}(x|z)] - \beta \text{KL}[q_{\phi_x}(z|x),p(z)] \nonumber \\
    & + \mathop{\mathbb{E}}_{q_{\phi_y}(z|y)}[\log p_{\theta_y}y|z)] - \beta \text{KL}[q_{\phi_y}(z|y),p(z)]
    \label{eqn:mmelbo}
\end{align}

To parameterize $q_{\phi_x}$ and $p_{\theta_x}$, we use architectures from \cite{bowman2015generating}. For $q_{\phi_y}$ and $p_{\theta_y}$, we use 3-layer MLPs\footnote{$q_{\phi_x}(z|x)$ is composed of  the program network and a stochastic layer; $p_{\theta_y}(y|z)$ is equivalent to the feedback network.}. To the best of our knowledge, this is the first application of a deep generative model to the feedback challenge.

\begin{figure*}[t]
    \includegraphics[width=\textwidth]{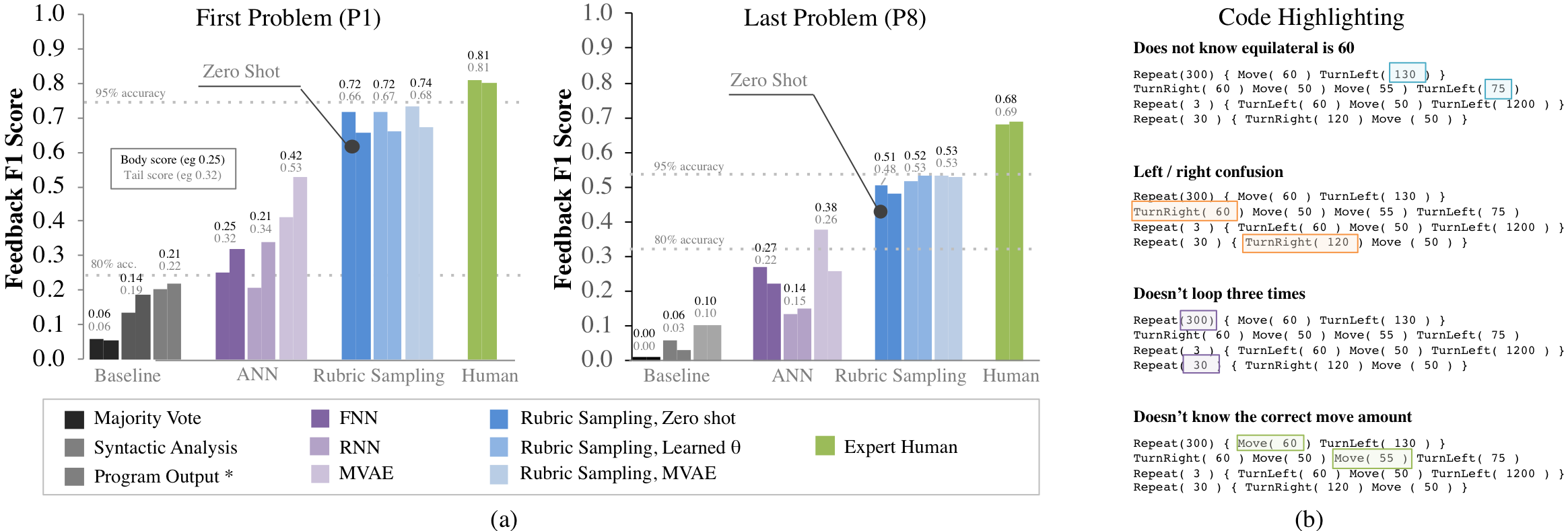}
\caption{(a) The F1 scores for P1 and P8. We plot two bars for each model representing the F1 score on the body (left) and on the tail (right). Rubric sampling models perform far better than baselines and grow close to human-level. The ``Zero Shot" marking refers to rubric sampling without fine-tuning. (b) Highlighting sub-programs conditioned on 4 feedback labels. The MVAE contains a modality for highlighting masks generated using the rubric. Imagine programming education where sections of a student's code can be highlighted along with helpful diagnostics.}
\label{fig:main_results}
\end{figure*}

\subsection{Rubric Sampling}
So far the models have been severely constrained by the number of labels. If we had a more efficient labeling strategy, we could better train these deep models to their full potential. For instance, imagine instead of focusing on individual programs, we ask experts to describe a student's thought process, enumerating strategies to get to a right or wrong answer. Given a detailed enough description, we can use it to label indefinitely. Granted, these labels will be noisy but the quantity should make up for any uncertainty. In fact, we can formalize such a ``description" as a \textit{context-free grammar}.

A context-free grammar (CFG) is composed of a set of acyclic production rules describing a space of output strings. As its name suggests, each rule is applied regardless of context (meaning no conditional arguments). Formally, a production rule is made up of \textit{non-terminal} and \textit{terminal} symbols. Non-terminal symbols are hidden and either produce another non-terminal symbol or a terminal one. Terminal symbols are made up of tokens that appear in the final output string. For instance, consider the following CFG: $\text{S} \rightarrow \text{AA}; \text{A} \rightarrow \alpha; \text{A} \rightarrow \beta$. $\text{S}$ and $\text{A}$ are non-terminal symbols whereas $\alpha$ and $\beta$ are terminal. It is easy to see that this CFG can only generate one of $\{\alpha\beta, \beta\alpha, \alpha\alpha, \beta\beta\}$. A \textit{probabilistic} context-free grammar (PCFG) is a CFG parameterized by a vector $\theta$ where each production rule has a probability $\theta_i$ attached. For example, we can make our CFG from above probabilistic: $\text{S} \xrightarrow{1.0} \text{AA}; \text{A} \xrightarrow{0.9} \alpha; \text{A} \xrightarrow{0.1} \beta$. Now, the space of possible outputs has not changed but for instance, $\alpha\beta$ will be more probable than $\beta\alpha$.

For the feedback challenge, the non-terminal symbols are labels, $y_i$ and the terminal symbols are programs, $x_i$. For example, a possible production rule might be:
\begin{equation*}
    \text{Correctly identified 45 degree angle} \xrightarrow{1.0} Turn(45)
\end{equation*}
With a PCFG, we can generate an infinite amount of synthetically labeled programs, $D_\text{syn} = \{x_i, y_i\}$, and use $D_\text{syn}$ to train data-hungry models. We refer to this as \textit{rubric sampling}. In practice, we sample a million synthetic examples. When training supervised networks, we only include unique examples to avoid prioritizing programs in the Zipf head.

Creating rubrics is surprisingly easy. For a novice (undergraduate) and an expert (professor), making a PCFG took 19.4 minutes. To make the process even easier, we developed a simple meta language for representing a grammar.\footnote{A Pytorch implementation, rubric grammars, along with data can be found at \url{https://github.com/mhw32/rubric-sampling-public}.}

\subsection{Further Learning from Unlabeled Programs}

As students use the platform, unlabeled submissions accumulate over time. We refer to the dataset as $D_{\text{unlabeled}}$.

\paragraph{Evolutionary Strategies} We can use unlabeled data in rubric sampling to automatically learn $\theta$. This means alleviating some of the burden for a human-in-the-loop, since choosing $\theta$ is often more difficult than designing the grammar itself. Intuitively, it is easier to reason about what mistakes a student can make than how often a student will make each mistake. But since a PCFG is discrete, we cannot  differentiate. However, we can hope to approximate local gradients by sampling  $\theta$ values within some $\epsilon$-neighborhood and computing finite differences along these random directions \cite{salimans2017evolution}. If we repeatedly take a linear combination of the ``best" samples as measured by a \textit{fitness} function, then over many iterations, we expect the PCFG to improve. The challenge is in picking the fitness function.

A good choice is to pick $\theta$ whose generated distribution, $D_\text{syn}$ is ``closest" to $D_\text{unlabeled}$\footnote{In tuning $\theta$, we consider all programs, not just the unique set (as doing so would not result in a sensible density estimator).}. As both are Zipf-ian, we can properly measure ``closeness" using a rank order metric \cite{havlin1995distance}, as rank is independent of frequency.

\paragraph{Rubric Sampling with MVAE}
Another way to service unlabeled data is to train with it: one of the features of the MVAE is that it can handle missing modalities. We can fit the MVAE with \textit{two} data sources: $D_\text{syn}$ and $D_\text{unlabeled}$.

In the case of missing labels, Equation~\ref{eqn:mmelbo} decomposes into the (unimodal) ELBO \cite{wu2018multimodal}:

\begin{equation}
    \mathop{\mathbb{E}}_{q_{\phi_x}(z|x)}[\log p_{\theta_x}(x|z)] - \beta \text{KL}[q_{\phi_x}(z|x),p(z)]
    \label{eqn:elbo}
\end{equation}

Thus, the MVAE is shown both a synthetic minibatch, $(x_i, y_i) \sim D_\text{syn}$, which is used to compute the multimodal elbo, and an unlabeled minibatch, $(x_i) \sim D_{\text{unlabeled}}$, which computes Equation~\ref{eqn:elbo}. We can jointly optimize the two losses by summing the gradients prior to taking an optimization step. Intuitively, this interpolates between $D_\text{unlabeled}$ and $D_\text{syn}$, no longer completely relying on the PCFG. One can also interpret this as a soft-version of structure learning since using $D_{\text{unlabeled}}$ is somewhat akin to ``editing" the PCFG.

\begin{figure}[h!]
    \centering
    \includegraphics[width=0.7\columnwidth]{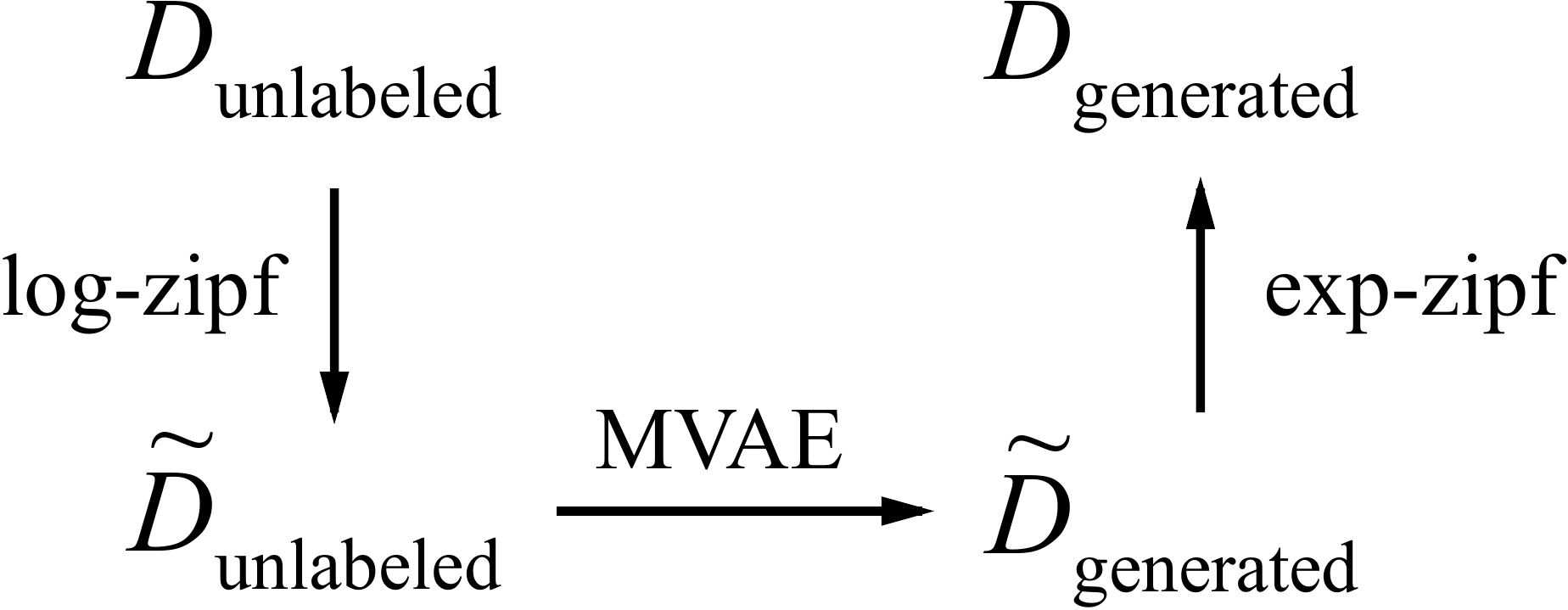}
    \caption{\textit{log-Zipf transformation}. Applying a log to frequencies of $x_i \sim D$ is like  ``tempering". This helps mitigate the effect of a few elements dominating the distribution.}
    \label{fig:log_zipf}
\end{figure}
\paragraph{Log-Zipf Transformation} Capturing a Zipf is hard for a generative model since it is very easy to memorize the top 5 programs and very hard to capture the tail. To make it easier, we apply a \textit{log} transformation,\footnote{We preserve examples that appear only once in $D$ to $\tilde{D}$ i.e. $\tilde{x} = \min(\log(x), 1)$ where $x \in D$ and $\tilde{x} \in \tilde{D}$.} e.g. if a program appears 10 times in $D$, it only appears once in the transformed dataset, $\tilde{D}$. Then, when generating with the MVAE, we invert the log by exponentiating the frequency of each unique program (exp-Zipf). Intuitively, log-Zipf is similar to ``tempering" a distribution as it reduces any extreme peaks and troughs.

\section{Results}
\subsection{Recreation of Human Labels}

Figure~\ref{fig:main_results} reports a set of F1 scores, including human-level performance estimated from annotations. Each model is given two bar plots, one for programs in the body (left) and one in the tail (right). First, we see that baselines have lower F1 scores compared to models that take advantage of synthetic data. That being said, the new baseline MVAE we introduced already performs far better than the previous SOTA. In P1, using rubric sampling increases the F1 score by 0.31 in the body and 0.13 in the tail (we find similar  gains in P8). These promising results imply that the grammar indeed is effective. We also find that combining the MVAE with rubric sampling boosts the F1 by an additional 0.2, reaching 94\% accuracy in P1 and 95\% in P8. With these scores, we are reasonably confident that for a new student, despite the likelihood that he/she will submit a never-before-seen program, we will provide good feedback.


To put these results in terms of impact, Table~\ref{tab:projections} estimates the number of correct feedback we could have given to students in P1 to P8 based on projections from the annotated set. Over the full curriculum, our best model would have provided the correct feedback to an expected 126,000 additional programs compared to what Code.org currently uses, potentially helping thousands more students.

\begin{table}[h]
    \centering
    \begin{tabular}{l|c}
        \toprule
        Model & \shortstack{Amount of \\ Correct Feedback} \\
        \hline
        Predicting from output & 1,483,157 (86.0\%) \\
        Rubric sampling with MVAE & \textbf{1,610,020} (93.7\%) \\
        Expert human & 1,658,162 (96.2\%) \\
        \bottomrule
    \end{tabular}
    \caption{\textit{Amount of correct feedback over the curriculum.} We ignore programs in the head of the Zipf as those can be manually labeled. With the best model, we could have provided 126,000 additional points of feedback.}
    \label{tab:projections}
\end{table}

\begin{figure*}[t!]
    \centering
    \includegraphics[width=\textwidth]{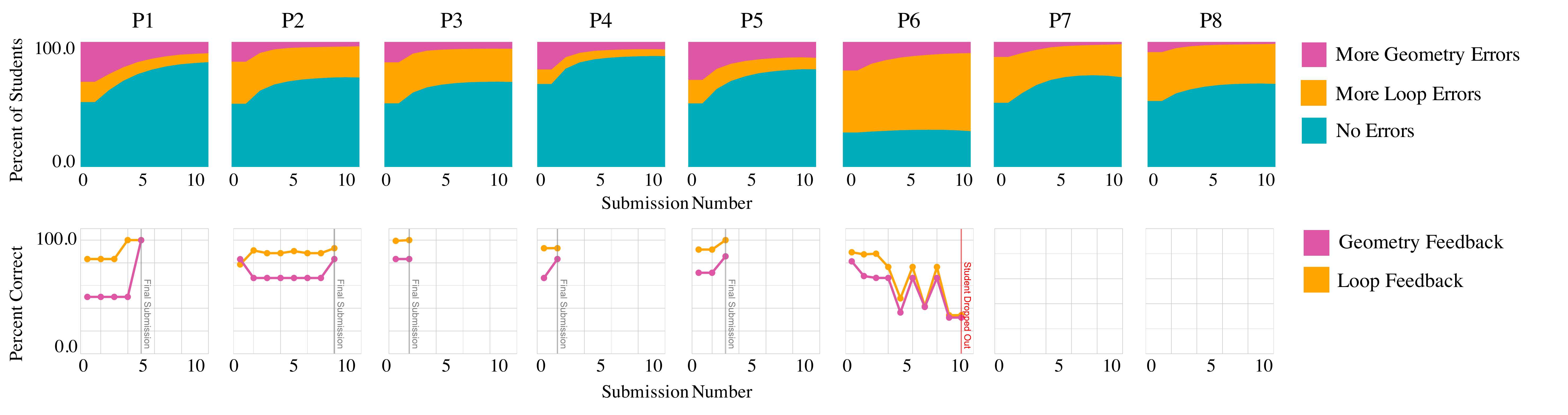}
    \caption{\textit{Student understanding of loops and geometry across curricula}: (top row) We plot the percentage of students who are either doing perfect (cyan), struggling more with looping concepts (orange), or struggling more with geometry concepts (pink). In general the percentage of students with no errors increases over time as more students finish the problem. Additionally, we can extrapolate that students are more effectively learning geometry than looping, as the area covered by pink decreases faster and more consistently than the area covered by orange. We can also see that P6 is somewhat of an outlier, being much more difficult for students than any other problem. (bottom row) In addition to aggregate statistics, we can also track learning for individual students. We can infer that this particular student tries several attempts with P6 before dropping out. }
    \label{fig:knowledge_over_time}
\end{figure*}

\subsection{Tracing Knowledge Across Curricula}
If we had a scalable way to provide feedback, what impact could we make to code education? We can begin to measure this impact using the full Code.org curriculum. Having demonstrated good performance on P1 and P8, we can confidently apply rubric sampling to P2 through P7. This is quite powerful as it allows us to estimate student understanding over a curricula. Critically, we can gauge the performance of both individual students and the student body as a whole. These sort of queries are valuable to teachers to be able to (1) measure a student's progress scalably and (2) judge how useful assignments and  lessons have been.

In Figure~\ref{fig:knowledge_over_time}, we analyze the average student's level of understanding over the 8 problems for two main concepts: loops and geometry (e.g. shapes, angles, movement). For each submission in a student's  trajectory, we classify it as having either 1) no errors, 2) more loop errors, or 3) more geometry errors\footnote{We do so by comparing the summed predicted probabilities for all labels related to loops, $\hat{y}_{i,\text{loop}} = \sum_{j \in \text{loop}} \hat{y}_{i,j}$ and labels related to geometry, $\hat{y}_{i,\text{geom}} = \sum_{j \in \text{geom}} \hat{y}_{i,j}$. If both $\hat{y}_{i,\text{loop}} < 1$ and $\hat{y}_{i,\text{geom}} < 1$, then we classify this program as ``no errors". Otherwise, we classify based on which quantity is larger.}. The figure shows the distribution of students in each of the three categories from the first 10 submissions. From looking at behavior within a problem and between problems, we can make the following inferences:

\begin{enumerate}
    \item \textbf{Most students are successfully completing each problem.} In other words, the blue area is increasing over time. Equivalently, the number of students still struggling by the 10th submission approaches a small number.
    \item \textbf{The difficulty of problems is not uniform.} P6 is much more difficult than the others as the fraction of students with correct solutions is roughly constant. In contrast, P1, P4, and P5 are easier, where students quickly cease to make mistakes. As a teacher, one could use this information to improve the educational content and better hone in on areas where more students struggle.
    \item \textbf{Students are learning geometry better than looping.} The rate that the pink area approaches zero is consistently faster than that of the orange area. By P8, students are barely struggling with geometry but a fair proportion still find looping difficult. As the curriculum was intended to teach nested looping, one interpretation is that the drawing aspect was somewhat distracting.
\end{enumerate}

\subsection{Fine-grain Feedback: Code Highlighting}

With most online programming education, the form of feedback is limited to pop-up tips or compiler messages. But, with generative models we can provide more fine-grain feedback through \textit{highlighting} subsets of the program responsible for the model predicting each feedback label.

First, if we consider a PCFG, the task of highlighting a program $x_i$ is equivalent to finding the most likely parsing in a probabilistic tree that would generate $x_i$. In practice, we use the A* algorithm for fast Viterbi parsing \cite{klein2003parsing}. Given the most likely parsing, we can follow the trajectory from root to leaf and record which sub-programs are generated by non-terminal nodes. This has one major limitation: only programs within the support of the PCFG can be highlighted. To bypass this, we can curate a synthetic dataset with each program having a segmentation mask denoting which tokens to highlight. If we treat the mask as an additional modality, we can then learn the joint distribution over programs, labels, and masks. See \cite{wu2018multimodal} for details in defining a VAE with three modalities. In Figure~\ref{fig:main_results}b, we randomly sample 4 programs and show segmentation masks. The running time to compute a mask is negligible, meaning that this can be used for providing on-the-fly feedback to students. Moreover, highlighting provides a notion of interpretability (which is extremely important if we are working with students), much like Grad-Cam \cite{selvaraju2017grad} did for computer vision.

\subsection{Clustering Students by Level of Understanding}
\begin{figure}[h!]
\centering
\includegraphics[width=\linewidth]{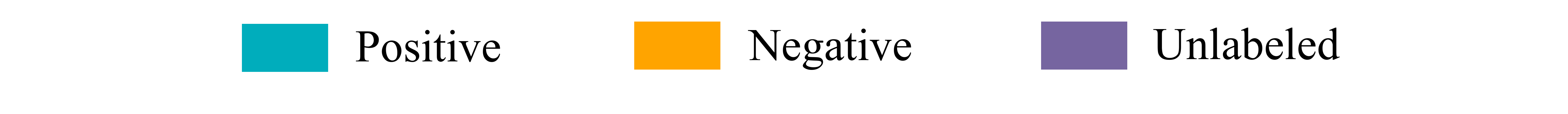}
\begin{subfigure}[t]{.32\columnwidth}
  \centering
  \includegraphics[width=\linewidth]{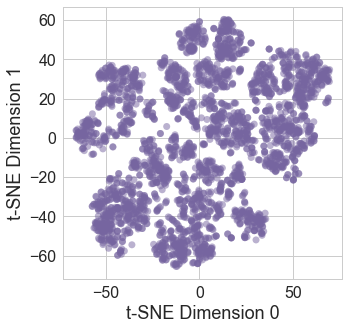}
  \caption{$D_\text{unlabeled}$}
\end{subfigure}%
\begin{subfigure}[t]{0.32\columnwidth}
    \centering
    \includegraphics[width=\linewidth]{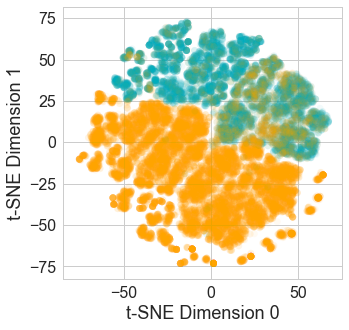}
    \caption{$D_\text{syn}$: Turn/Move}
\end{subfigure}
\begin{subfigure}[t]{0.32\columnwidth}
    \centering
    \includegraphics[width=\linewidth]{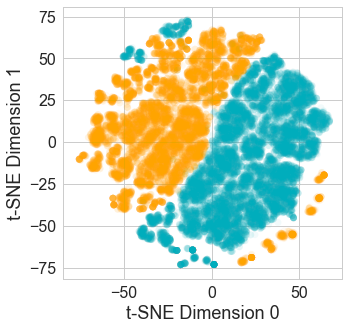}
    \caption{$D_\text{syn}$: No Repeat}
\end{subfigure}
\caption{\textit{Clustering students.} Using the inference network in the MVAE, we can embed a program in 2D. In $D_\text{unlabeled}$ (a), we see a handful of distinct clusters. In $D_\text{syn}$ (b,c), we find meaningful clusters that are segmented by labels.}
\label{fig:clusters}
\end{figure}

With any latent generative model, the rich latent space provides a vector representation for unstructured data. Using the MVAE, we first sample $z_i \sim q(z_i|x_i)$ for all $x_i \in D_{\text{test}}$; we then train use t-SNE  \cite{maaten2008visualizing} to reduce to two dimensions. In Figure~\ref{fig:clusters}b and c, we color each embedded program from $D_\text{syn}$ by whether the true label is positive or negative. We find that the space is partitioned to group programs with similar feedback together. In Figure~\ref{fig:clusters}a, we see that $D_{\text{unlabeled}}$ is also organized into disjoint clusters. This implies that even with no data about a new student we can draw inferences from their neighbors in latent space.

\section{Discussion}
\paragraph{We get closer to human level performance than previous SOTA.} Any of the rubric sampling models beat the SOTA by at least 0.46 in P1 and 0.24 in P8, nearly tripling the F1 score in P1 and doubling in P8. In both exercises, our best model is just under 95\% accuracy, which is encouraging for this to be implemented in the real world.

\paragraph{We can effectively track student growth over time.} With a high performing model, we can analyze students over time. For Code.org, we were able to (1) gauge what individual students struggle with, (2) gauge what a collective of students struggle with, (3) identify the effectiveness of a curriculum, and (4) identify the difficulty of problems.

\paragraph{Making a rubric is neither difficult nor costly.} It is not the case that only experts can make a good rubric. We also asked an undergraduate (novice) to make a rubric for P1 and found that while an experts' rubric averaged $0.69 \pm 0.03$ in F1 score, the novice's averaged $0.63 \pm 0.01$, both of which are much higher than baselines ($0.29 \pm, 0.03$). Furthermore, we measured that it took a group of teaching assistants 24.9 hours to label 800 unique programs. In comparison, it took a novice an average of 19.4 minutes to make a rubric.

\paragraph{We provide feedback to programs that do not compile.}
Rubic sampling and MVAE make no assumptions on program behavior, working out-of-the-box from the 1st student.

\paragraph{We do not need to handpick $\theta$ when designing a rubric.}

In Figure~\ref{fig:policy}, the PCFG uses  hand-picked $\theta$. However, one can argue that it is difficult to know how often students make mistakes and yet, the choice of $\theta$ is important: performance drops if we randomly choose. For example, in P1, using hand-picked $\theta$ has a $0.26 \pm 0.002$ increase over random $\theta$ in F1-score. In the absence of an expert, we can use evolutionary strategies to find a good minima starting from a random initialization. Over 3 runs, we found that learning $\theta$ reduces the difference to $0.007 \pm 0.005$ in P1, even beating expert parameters by $0.005 \pm 0.009$ in P8. With a large dataset, we only have to define the structure, not the probabilities.

\paragraph{The log-Zipf transform ensures that learning does not collapse to memorizing the most frequent programs.} If we were to use the raw statistics of the student data, the MVAE would minimize loss by memorizing the most frequent programs. As evidence of this, when sampling from its generative model, we found that the MVAE only generated programs in the top 25 programs by frequency (even with 1 million samples). We propose the log-Zipf transformation as a general tool for parameter learning with Zipf-distributed data. Critically, the transform downweights the head and upweights the tail in an invertible fashion.

\paragraph{The MVAE interpolates between synthetic and empirical data.} Unlike the PCFG, we can train the MVAE with multiple data sources. Effectively, the programs we show to the model is drawn from an interpolation between the synthetic distribution defined by rubric sampling and the true (unlabeled) data distribution. Intuitively, the increase in performance from the MVAE can be attributed to better capturing the true density of student programs (see Figure~\ref{fig:limitations}).

\begin{figure}[h!]
    \begin{subfigure}[h]{0.49\columnwidth}
        \centering
        \includegraphics[width=\linewidth]{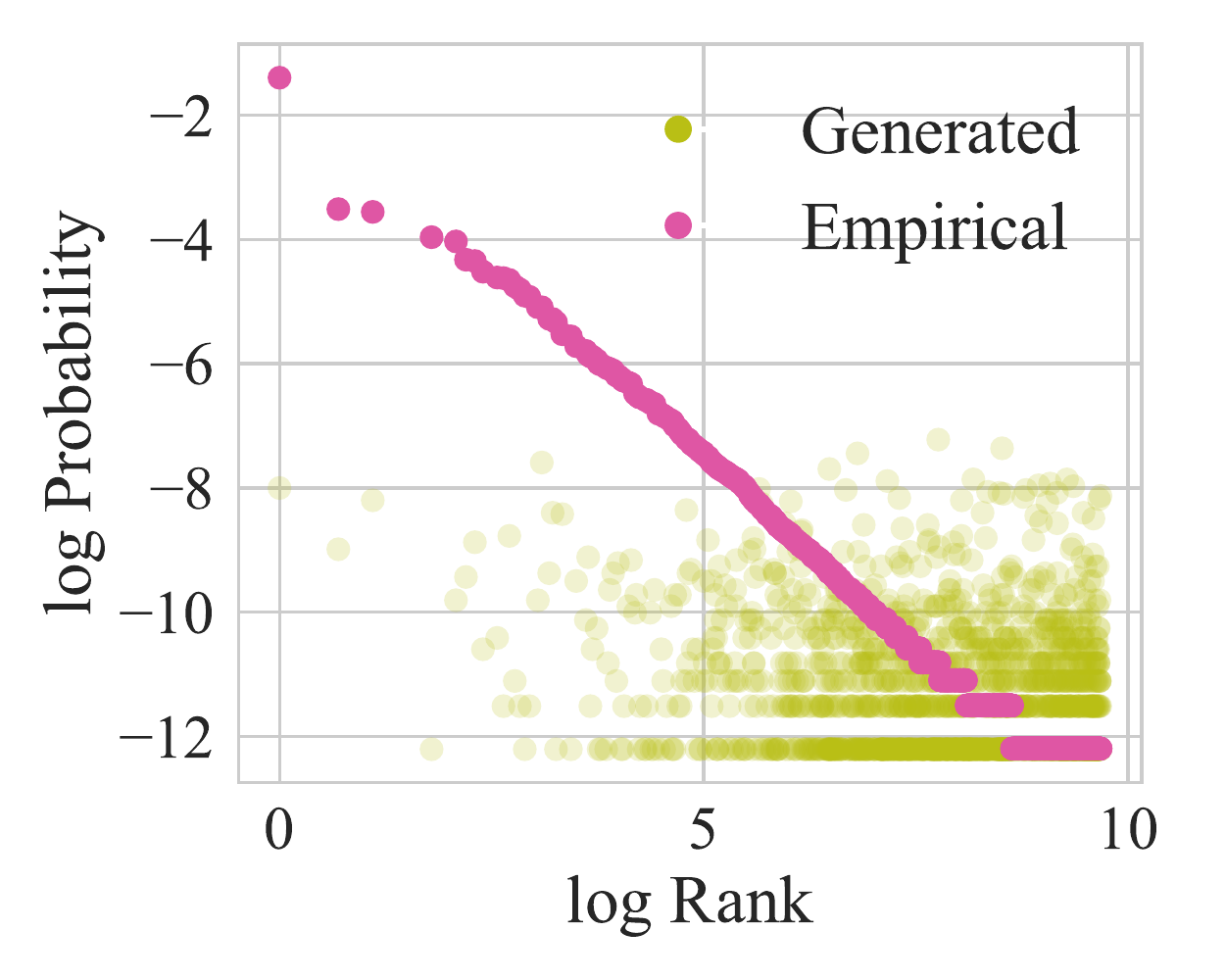}
        \caption{P1 (PCFG)}
    \end{subfigure}
    \begin{subfigure}[h]{0.49\columnwidth}
        \centering
        \includegraphics[width=\linewidth]{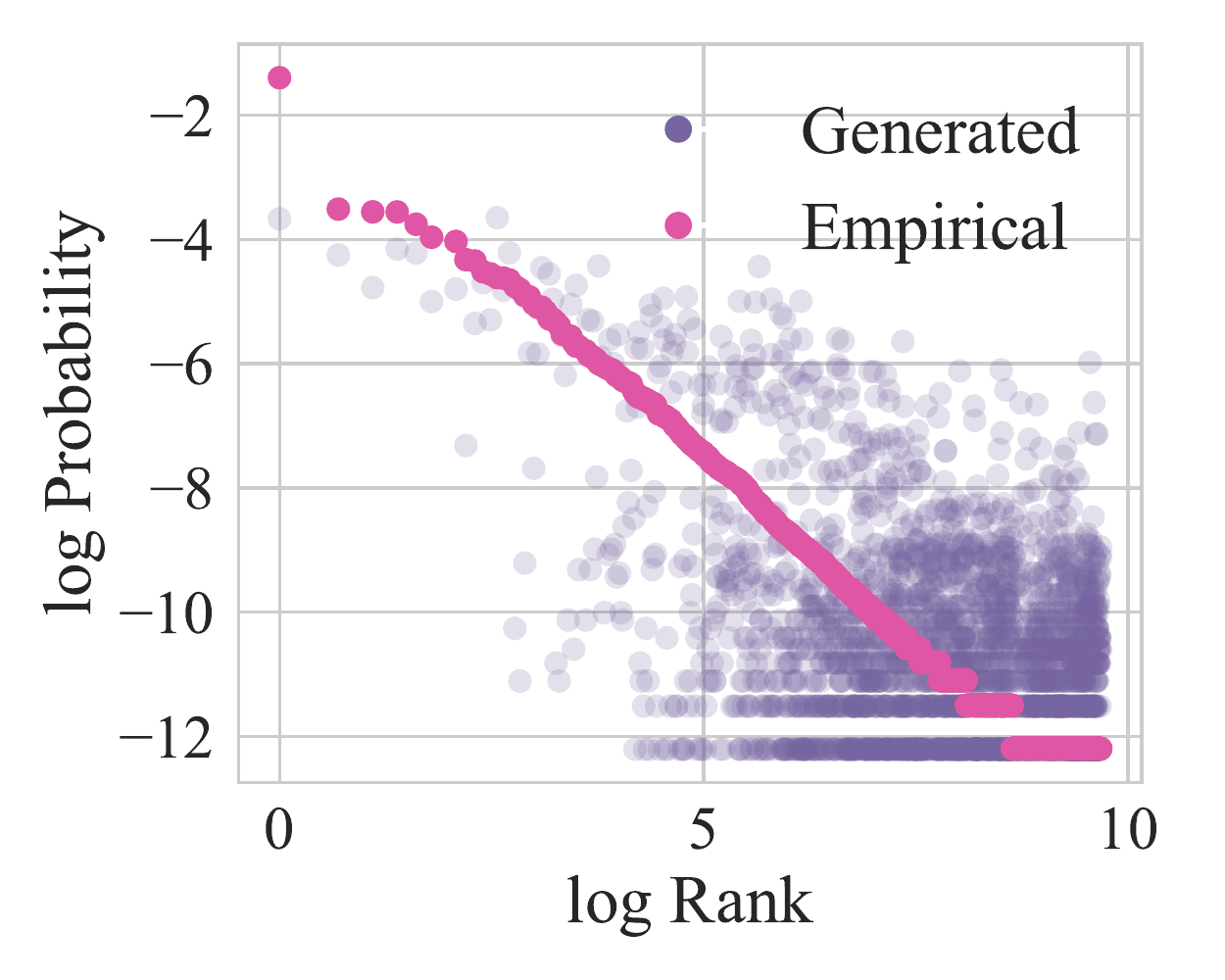}
        \caption{P1 (MVAE)}
    \end{subfigure}
    \caption{We compare $D_\text{syn}^{MVAE}$ and $D_\text{syn}^{Rubric}$ to $D_\text{unlabeled}$. Programs from the MVAE cover $D_\text{unlabeled}$ much better than relying on synthetic data alone (PCFG). }
    \label{fig:zipf_outputs}
\end{figure}

\paragraph{We can generate and open-source a large dataset of student programs.} Datasets with student data are difficult to release to the open community. But large public datasets have been a dominant force in pushing the boundaries of research. Luckily, with generative models, we can curate our own ``Imagenet" for education. But, we want to ensure that our dataset matches $D$ in distribution. Intuitively, it is impossible that a PCFG can capture $D$ since that would require production rules that span the entire tail of the Zipf. In fact, as shown in Figure~\ref{fig:zipf_outputs}, the PCFG is not that faithful.  One remedy is to use the MVAE trained with $D_{\text{unlabeled}}$ as that is interpolating between distributions. Figure~\ref{fig:zipf_outputs}, confirms that the MVAE matches the shape of $D$ much better.

\section{Limitations and Future Work}
\begin{figure}[h!]
\centering
\includegraphics[width=0.7\linewidth]{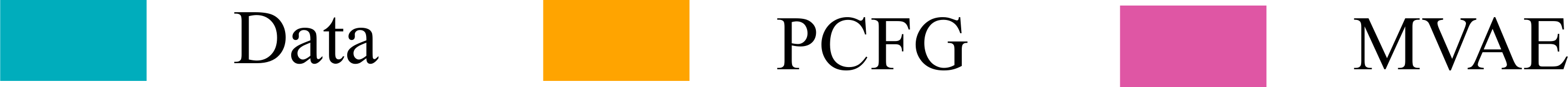}\\\vspace{0.25cm}
\begin{subfigure}[t]{.49\columnwidth}
  \centering
  \includegraphics[width=0.55\linewidth]{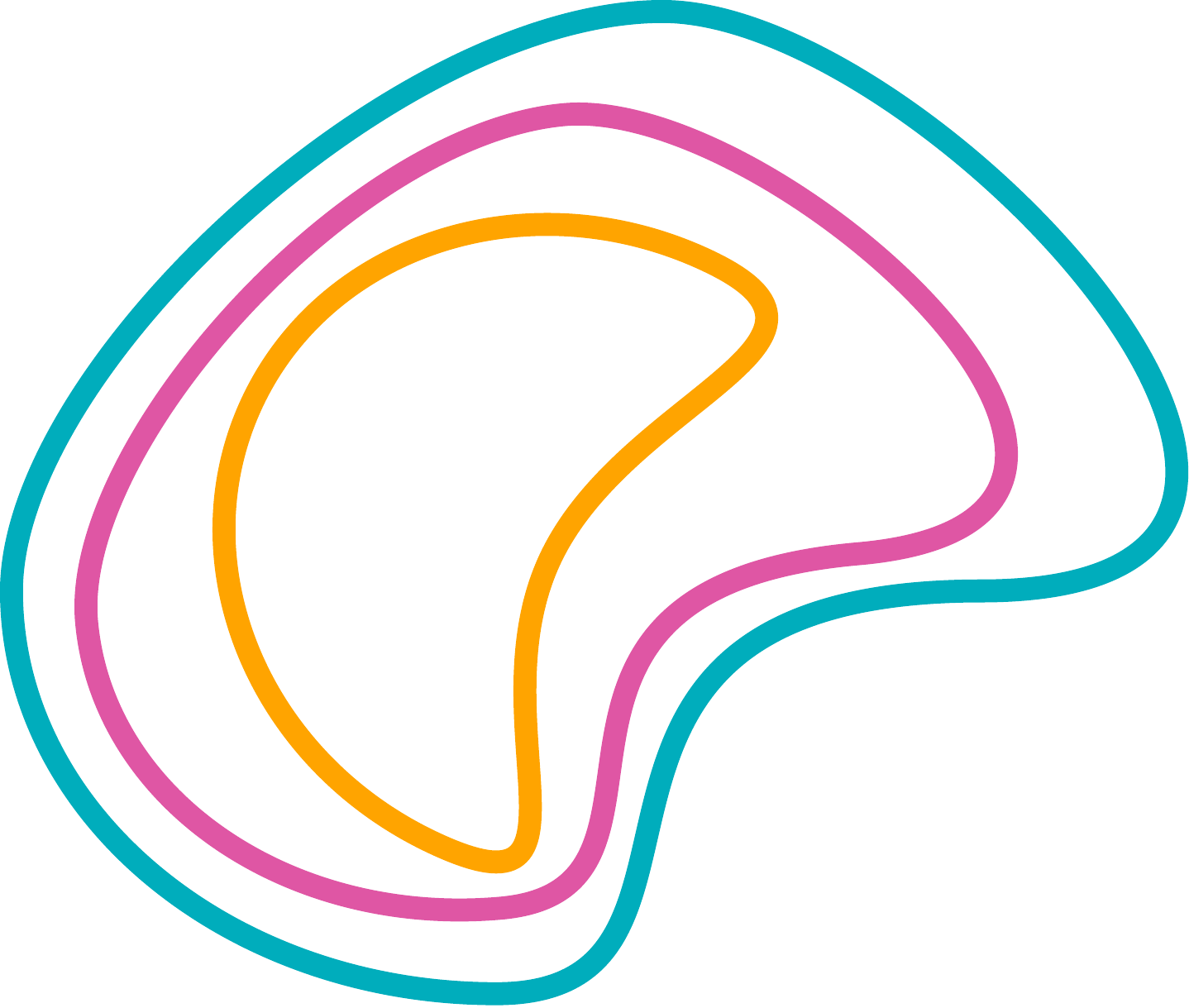}
  \caption{Code.org}
\end{subfigure}%
\begin{subfigure}[t]{0.49\columnwidth}
    \centering
    \includegraphics[width=0.55\linewidth]{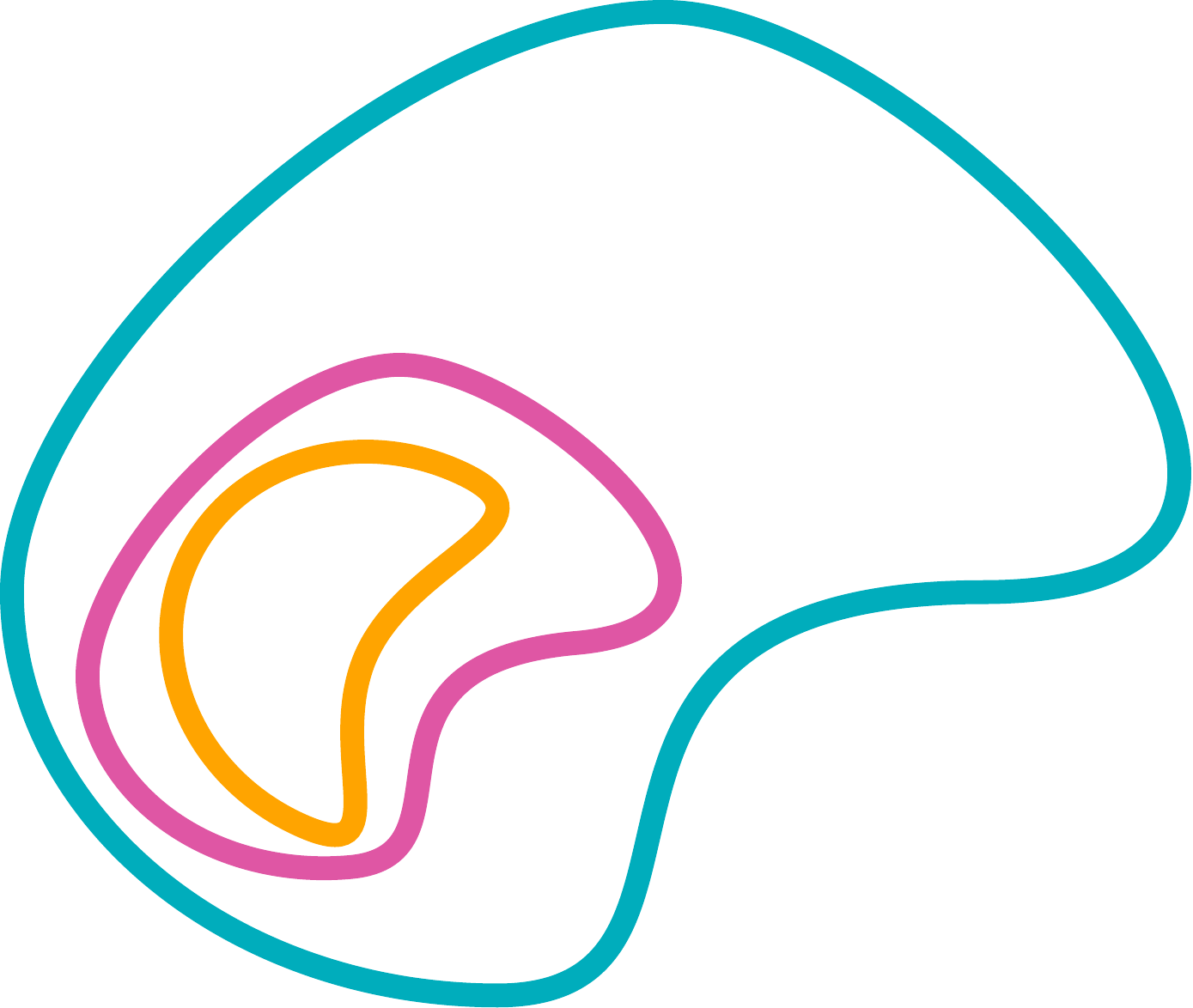}
    \caption{University classroom}
\end{subfigure}
\caption{The space of likely student programs in a block-based language can be covered by a PCFG. But in a higher-order language like Python, the space is much larger.}
\label{fig:limitations}
\end{figure}

The effectiveness of rubric sampling is largely determined by the complexity of the programming task. Although a block-based language like in Code.org already has an infinite number of possible programs, the set of probable student programs is much smaller than in a typical university level coding assignment. An  important distinction is the introduction of variable and function names. As suggested by Figure~\ref{fig:limitations}, the version of rubric sampling used in this paper may have difficulty scaling to harder problems. As PCFGs are context-free, making a sufficiently expressive grammar for complex problems requires extremely large graphs with repetitive subgraphs. Future work could look to generalizing rubric sampling to arbitrary graphs with conditional branching, or investigate structural learning to improve graphs to cover more of the empirical data.

\section{Conclusion}
We introduce the zero shot feedback challenge. On a widely used platform, we show rubric sampling to far surpass the SOTA. We combine this with a generative model to cluster students, highlight code, and incorporate historical data. This approach can scale feedback for real world use.

\section{Acknowledgments}
The authors thank Lisa Wang, Baker Franke and the Code.org team for support, and guidance. MW is supported by NSF GRFP. NDG is supported by DARPA PPAML under FA8750-14-2-0006.  We thank Connie Liu for edits.

\bibliographystyle{aaai}
\bibliography{zeroshot}

\end{document}